\documentclass[num,sort,preprint,review,11.5pt]{elsarticle}

\usepackage{graphicx}
\usepackage{textcomp}
\usepackage{colortbl,xcolor}
\usepackage{float}
\usepackage{adjustbox}
\usepackage{hyperref}
\usepackage{amsmath}
\usepackage{booktabs}
\usepackage{hhline}
\usepackage{orcidlink}
\usepackage{algorithm}
\usepackage{amsthm}
\usepackage[noend]{algpseudocode}
\usepackage{amsfonts}
\usepackage{caption}
\usepackage{amssymb}
\algnewcommand{\algorithmicand}{\textbf{ and }}
\algnewcommand{\AND}{\algorithmicand}

\usepackage{algorithm}
\usepackage[noend]{algpseudocode}
\makeatletter
\renewcommand{\ALG@beginalgorithmic}{\tiny}
\algrenewcommand\alglinenumber[1]{\tiny #1:}
\makeatother
\newtheorem{definition}{Definition}

\begin{document}

\begin{frontmatter}



\title{Extracting Process-Aware Decision Models from Object-Centric Process Data}


 \author{Alexandre Goossens\orcidlink{0000-0001-8907-330X},
 Johannes De Smedt \orcidlink{0000-0003-0389-0275},
 Jan Vanthienen\orcidlink{0000-0002-3867-7055}}

 \affiliation{organization={Leuven Institute for Research on Information Systems (LIRIS), KU Leuven},
             addressline={Naamsestraat 69}, 
             city={Leuven},
             postcode={3000}, 
             country={Belgium},
             }
 \ead{alexandre.goossens@kuleuven.be, johannes.desmedt@kuleuven.be, jan.vanthienen@kuleuven.be}

\begin{abstract}
Organizations execute decisions within business processes on a daily basis whilst having to take into account multiple stakeholders who might require multiple point of views of the same process. Moreover, the complexity of the information systems running these business processes is generally  high as they are linked to databases storing all the relevant data and aspects of the processes.
Given the presence of multiple objects within an information system which support the processes in their enactment, decisions are naturally influenced by both these perspectives, logged in object-centric process logs.
However, the discovery of such decisions from object-centric process logs is not straightforward as it requires to correctly link the involved objects whilst considering the sequential constraints that business processes impose as well as correctly discovering what a decision actually does. This paper proposes the first object-centric decision-mining algorithm called Integrated Object-centric Decision Discovery Algorithm (IODDA). IODDA is able to discover how a decision is structured as well as how a decision is made. Moreover, IODDA is able to discover which activities and object types are involved in the decision-making process. Next, IODDA is demonstrated with the first artificial knowledge-intensive process logs whose log generators are provided to the research community. 
\end{abstract}




\end{frontmatter}


\section{Introduction} \label{Introduction}

Organizations and companies regularly execute business processes to achieve specific objectives, like delivering a product or creating a website. 
Business processes can be monitored with so-called process-aware information systems logging each step of the process hence storing the history of business processes in process logs.
Traditionally, business processes have been logged and discovered from a single point of view with, e.g., only the traveler's or the airplane's point of view in an airport luggage handling process  \cite{li2018extracting,ghahfarokhi2020ocel}. 
However, business processes, like information systems, often include multiple object types within their process, e.g., in a luggage handling process objects include travelers, luggages, airlines, airplanes and the airport, with distinct perspectives and interactions (e.g., passenger checking in luggage with an airline) in the system.
This complexity means that traditional single object type process logs,
such as eXtensible Event Stream (XES) \cite{xes2014ieee}, are not always suitable to log and discover complete process models involving multiple object types \cite{van2019object}.
These so-called object-centric processes have recently gained traction with multiple object-centric process modelling proposals \cite{van2001proclets, kleijn2012regions,van2020discovering,ghilardi2022petri} and object-centric process logging proposals \cite{li2018extracting,van2017object,ghahfarokhi2020ocel,goossens2023enhancing} being introduced.
Object-centric event logs enable the logging and analysis of object-centric business processes.
For instance, in the airport luggage handling process, the check-in event can now be associated with all relevant objects like the passenger, luggage, and airplane, instead of just one. 
Furthermore, companies make decisions based on information from multiple object types, such as luggage handling decisions influenced by factors like the customer's ticket, luggage details (weight, dimensions and content), aircraft and potentially airport policies \cite{vanthienen2021decisions}. 
Decisions can therefore be considered as inherently object-centric and are fit to be logged alongside object-centric processes with object-centric process logs.

This study is particularly interested in knowledge-intensive processes (KiPs) where the order of activities is heavily influenced by the knowledge and decision-making of the process performers. 
Such KiPs are mainly driven by decisions which based on various inputs determine an output which in turn might influence individual process instances. 
Finally, it is considered good practice to model operational decisions in decision models \cite{vanthienen2021decisions}. A common standard for this is the Decision Model and Notation (DMN) standard allowing to model both the decision structure and decision logic \cite{DMN2015} allowing to communicate, model, execute and automate operational decisions. Note that DMN allows also for more complex rules involving multiple variables.

This paper presents a novel object-centric algorithm, the Integrated Object-centric Decision Discovery Algorithm (IODDA) which discovers both the decision structure and the decision rules present in the object-centric process log and brings both together in a complete DMN model \cite{DMN2015}. IODDA is also able to identify how and which activities and object types are involved in the decision-making process. The algorithm draws inspiration from a single-object perspective DMN discovery algorithm implemented on XES logs \cite{de2019holistic}. Due to a lack of publicly available knowledge-intensive object-centric process logs, this paper also provides the two first artificial knowledge-intensive process logs and log generators in the recent DOCEL standard.  The algorithm's effectiveness is validated through its successful application on these two artificial logs, effectively rediscovering the relationships between variables and the underlying logic governing process execution. The main contributions of this paper are:

\begin{itemize}
    \item Proposing the first algorithm, IODDA, to discover object-aware DMN models from Data-aware Object-Centric Event Logs (DOCEL). 
    \item The first artificial DOCEL log generators for two different problems are made available together with the set of artificial DOCEL logs used in this paper.  The source code of the log generators is also provided so that the research community can replicate this for other knowledge-intensive process generators. 
\end{itemize}

The rest of the paper is structured as follows: Section \ref{Related Work} deals with the related work followed by Section \ref{Problem Statement} illustrates the problem. Next, Section \ref{Preliminaries} deals with the preliminaries and Section \ref{P-MinD Algorithm} introduces the novel algorithm. Section \ref{Artificial DOCEL logs} introduces the artificial log generators with Section \ref{P-MinD in Action} illustrating how the algorithm works in practice. Sections \ref{Discussion} and \ref{Limitations and Future Work} respectively deal with the discussion, limitations and future work. Lastly, Section \ref{Conclusion} concludes this paper.

\section{Related Work} \label{Related Work}

The following section first deals with the field of object-centric processes and secondly an overview of works studying the extraction of decision models from single-object point of view process logs is provided.

\subsection{Current Research in Object-Centric Processes}
Despite the recent surge in interest for object-centric processes, the concept of multiple object types influencing a business process was already proposed in 2001 with the introduction of so-called proclets, which are sub-processes dealing with an artifact within the context of a business process \cite{van2001proclets}. Over the years, various object-centric visualization proposals have been introduced such as Colored Petri Nets (CPN) \cite{kleijn2012regions}, Catalogue and Object-aware Nets (COA) \cite{ghilardi2022petri} or Guard-Stage-Milestone models (GSM) \cite{hull2011business}. Despite these proposals, there is still no established standard modelling language for object-centric processes.

The execution of object-centric processes can be logged in event logs. However, due to the object-centric nature of these processes, a single object point of view event log such as XES logs results in some information loss due to divergence and convergence issues \cite{van2019object}. Divergence issues cause certain activities to not clearly be allocated to the correct object anymore whilst convergence issues incorrectly multiply certain activities to incorrectly match their actual number of occurrences.

Because an object-centric process can be viewed from the point of view of multiple objects simultaneously, the traditional concept of cases and traces is not valid anymore for object-centric processes. To execute an object-centric process it is therefore important to know which objects are related to one another and which objects are involved in each activity \cite{ghahfarokhi2020ocel, li2018extracting,goossens2023enhancing,galanti2023object}.

With the XES event logging format \cite{xes2014ieee} being unusable to store object-centric event processes \cite{van2019object}, other logging formats were introduced to solve this issue. Both Object-Centric Behavioral Constraint (OCBC) models \cite{van2017object} and eXtensible Object-Centric (XOC) logs \cite{li2018extracting} realized the importance of object-object relations when dealing with object-centric processes. As such, these event log formats would store the relevant object-object relations with each event together with all their attributes having the side-effect of causing scalability issues with these formats. To solve these scalability issues, Object-Centric Event Logs (OCEL) were introduced \cite{ghahfarokhi2020ocel}. An OCEL log stores the various objects and their attributes separately from the business process events whilst linking objects to events with the use of foreign keys. In \cite{rebmann2022uncovering}, the authors propose an algorithm to convert XES logs to OCEL logs. Even though OCEL addressed the issue of scalability, certain aspects regarding object-attribute allocation were not implemented due to design choices.  For example, in OCEL, it is not possible to unequivocally know which attribute belongs to which object when stored in the events table, making attribute value changes or even batch attribute value changes difficult to track \cite{goossens2023enhancing}. These issues have been addressed with the proposal of a Data-Aware Object Centric Event Log (DOCEL) where a distinction is made between static and dynamic attributes \cite{goossens2023enhancing}. Static attributes are attributes belonging to an object that do not change over the course of a process execution whilst dynamic attributes are attributes that can change value over the course of the process execution and are assigned to both an event and an object \cite{goossens2023enhancing}. In \cite{goossens2023ocel}, an algorithm is proposed to convert OCEL to DOCEL.

With object-centric processes having multiple object types, new precision and fitness conformance metrics were proposed in \cite{adams2021precision} and process performance measures in \cite{estrada2023defining}. With the concept of cases and traces of traditional process mining not being valid anymore for object-centric process mining, the study in \cite{adams2022defining} uses graph isomorphism to introduce a similar case concept usable for object-centric processes. In \cite{adams2022framework}, a framework to encode object-centric features using tabular, sequential and graph-based encodings is introduced.
Next, an approach to filter and sample object-centric event logs is laid out in \cite{berti2022filtering}. Object-centric process discovery was analyzed to discover Object-Centric Petri Nets \cite{van2020discovering} and Object-Centric Directly-Follows Multigraphs \cite{berti2022oc} respectively. To reduce the complexity of the discovered process models, the study in \cite{ghahfarokhi2022clustering} clusters similar objects in the object-centric event logs whilst the study conducted in \cite{delias2023formulating} also identifies interacting objects of different object types. 
In summary, the study in \cite{galanti2023object} shows that incorporating object-interactions into object-centric predictive analysis improves performance, highlighting the importance of adapting single object type algorithms for this context.

The discovery of decisions from object-centric event logs is not straightforward as it requires correctly matching activities with decisions as well as correctly identifying which attributes are used within the decision-making process. Next to that, it is crucial that the correct objects and object types are linked together. Finally, all this information is spread over multiple events which need to be analyzed in a control-flow conforming manner. In short, even though the variables are linked to the objects, this increase in dimensions also increases the complexity of the problem.

Currently, the discovery of decisions and their logic within object-centric processes have not been researched yet.

\subsection{Decision mining with traditional process logs}

Decision mining with single point of view logs has been researched before with the emphasis on decision point analysis. Decision point analysis aims at discovering gateways in a process where the routing of a case depends on the outcome of a certain decision \cite{rozinat2006decision}. Discovering complete DMN models with multiple decision levels is different though and implies discovering when decision variables are introduced or derived in the process and how these relate to one another. This analysis reduces the complexity of the process model as the decision logic is now separated from the process logic \cite{biard2015separation}. Works that focus on both control flow and decision analysis simultaneously to discover DMN models are not very prevalent.
In \cite{bazhenova2016discovering}, an approach to find DMN models from traditional event logs is proposed using decision trees. However, the approach is not capable of dealing with loops, nor can the discovered DMN model be linked to the business process making that the control flow and decision logic are still intertwined which is in conflict with the principle of separation of concerns\cite{biard2015separation}. 
 A DMN discovery algorithm has been proposed in \cite{de2019holistic} which is compliant with the principle of separation of concerns.
The previous algorithms were developed for XES logs and are not directly usable within an object-centric context.
Lastly, a broader overview of the decision mining field and its future from other sources such as textual descriptions can be found in the following studies  ~\cite{leewis2020putting,vanthienen2021decisions,de2019holistic}.

An important remark is that due to the dimension reduction taking place from object-centric logs to single object point of view logs, there is a loss of information taking place such that certain object relations might be lost or certain activity orders are not correct anymore making that the complete decision might not be discoverable anymore due to the convergence and divergence issues where for example a luggage might not be traced back to the correct traveler anymore or that it might be unclear on which luggage was put in first in an airplane.
The other way around where single object point of view logs are combined to create object-centric process logs will not generate more information given that the dimension reduction already took place and as such already reduced the available information.
Another point of view is that if a decision contains multiple objects, these need to be stored together to be analyzed together.
Therefore, there is a need for a native object-centric decision discovery algorithm that is developed for an object-centric event log format.

\section{Problem Illustration} \label{Problem Statement}

This section introduces a knowledge-intensive process dealing with a publication process to illustrate the problem. Next, the research goals are laid out.

\subsection{Running Example}

\begin{figure*}[h]
\centering
\includegraphics[width=.9\linewidth]{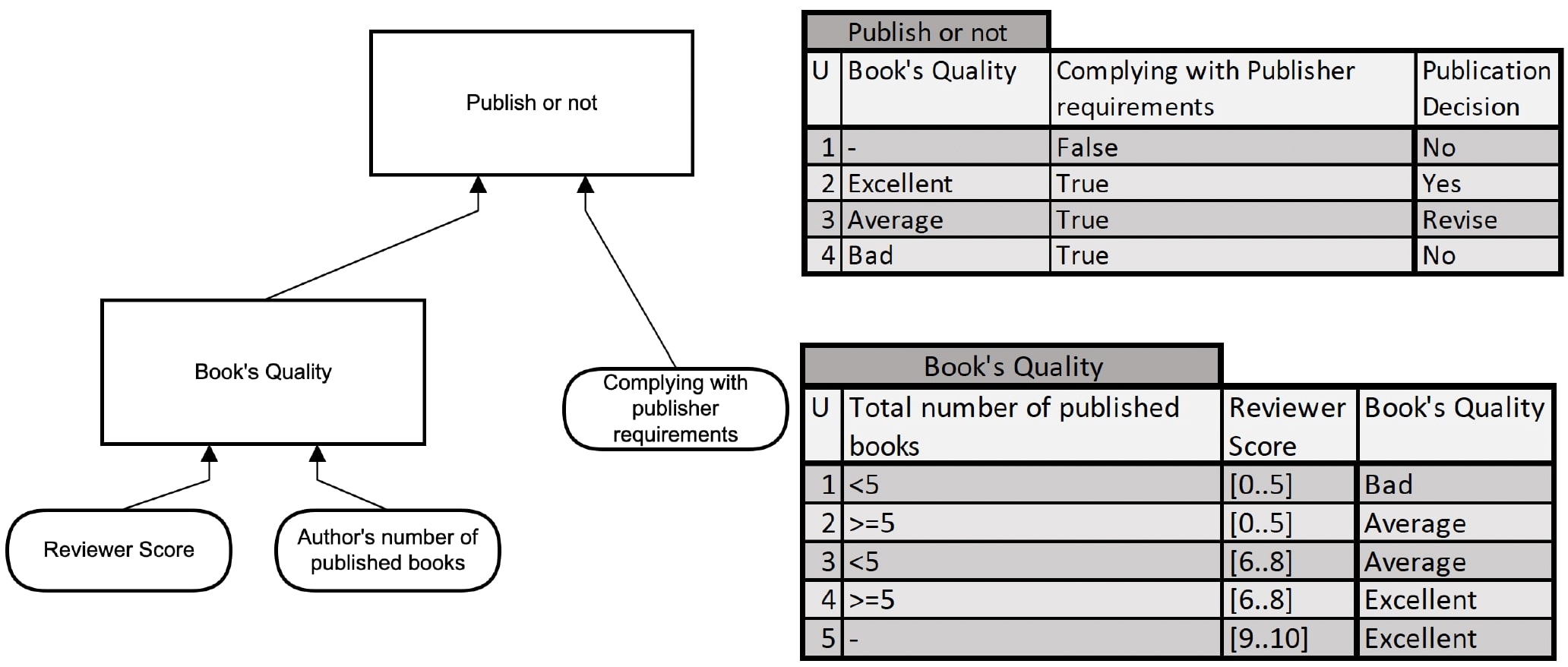}
\caption{DMN model of the running example} \label{DMNbookrunning}
\end{figure*}

\begin{figure*}[h]
\centering
\includegraphics[width=0.85\linewidth]{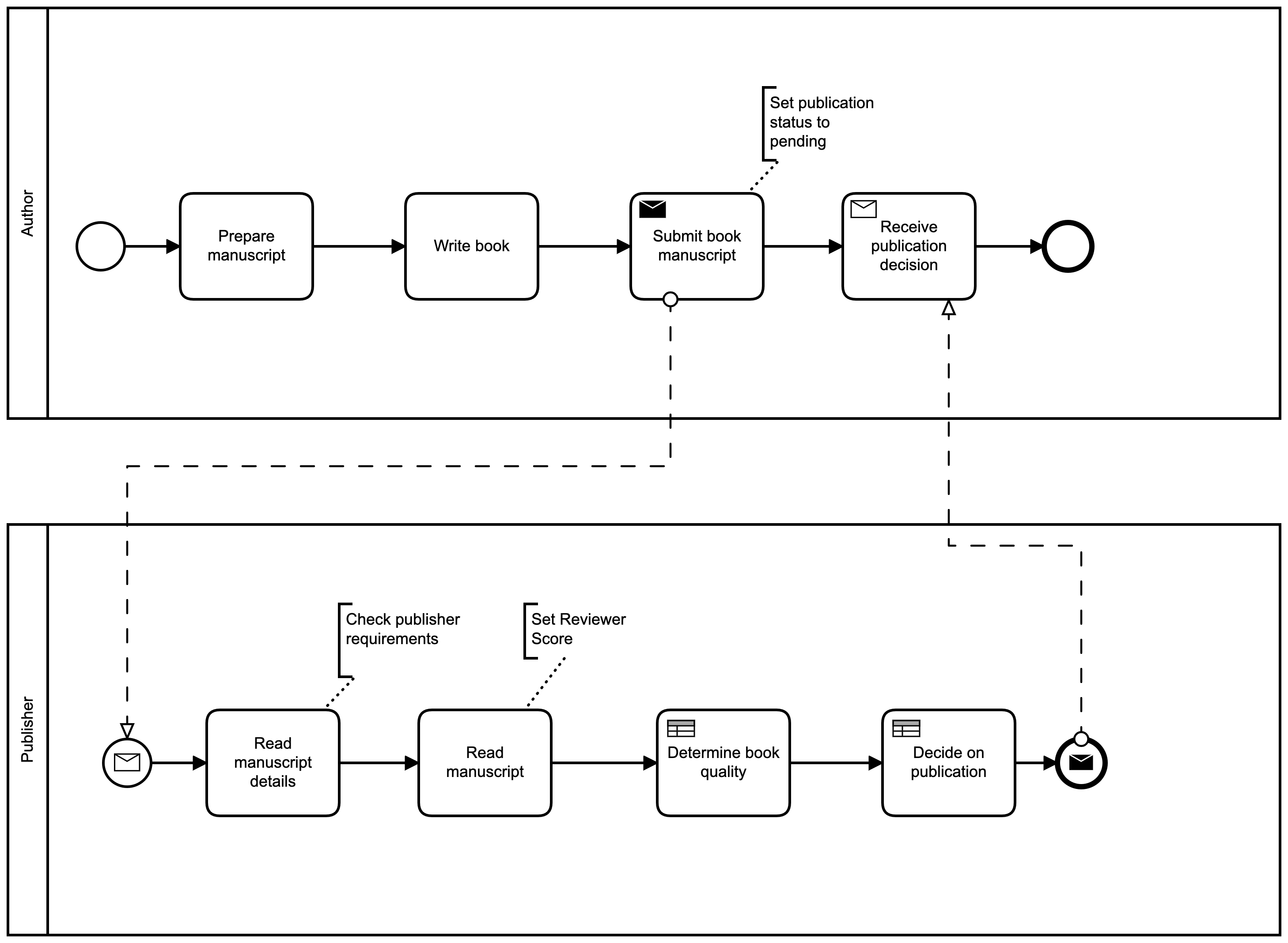}
\caption{BPMN model of the running example} \label{BPMNbookrunning}
\end{figure*}

The problem is illustrated with a knowledge-intensive publication process of a book. The DMN model driving the business process is shown in Figure \ref{DMNbookrunning} and the BPMN model is shown in Figure \ref{BPMNbookrunning}. The process is considered as the representation of the system generating object-centric event logs in which various decisions are embedded. Note that this first running example has a linear control flow for simplicity, but the second running example contains more gateways and loops. The first running example contains three different object types: \texttt{Author}, \texttt{Book} and \texttt{Publisher} with the following attributes:
\begin{itemize}
    \item \texttt{Author}: \texttt{Name, Author Specialty Genre, Total number of published books}
    \item \texttt{Book}: \texttt{Genre, Number of pages, Publication status, Review Score, Quality, Compliance with the publisher requirements}
    \item \texttt{Publisher}: \texttt{Name, Publisher Specialty Genre}
\end{itemize}

\textbf{Book publication process and decision}: \textit{Each author has already published a number of books in the past. This is considered to be available information at the start of the process. The process is initialized by an author who first has to \texttt{prepare a manuscript} and then \texttt{writes a book}. Afterwards, the author \texttt{submits their book manuscript} to a publisher which sets the publication status of that book to pending. Next, the book manuscript is received by the publisher and the publisher will read the manuscript details. The activity \texttt{Read manuscript details} will check whether the book complies with the publisher requirements. Next, this process takes the assumption that every submitted book is entitled to be read in more detail after which a reviewer score is given at the end of activity \texttt{Read manuscript}. The DRD (a Decision Requirements Diagram visualizes the general structure of a decision, see Section \ref{Preliminaries} for a more detailed explanation) of Figure \ref{DMNbookrunning} shows that a Book's Quality is determined by the reviewer score and by the amount of books an author has published in the past. For example, the decision table of Book's Quality shows that if an author has published 5 books in the past and the reviewer gives a score of 4 then the Book's Quality is considered to be average. This decision is executed by activity \texttt{Determine book quality}. Finally, a publication decision is made by the publisher by taking into account the Book's Quality and the compliance of publisher requirements in activity \texttt{Decide on publication}. This is then communicated to the author which ends the process.}

For this study, DOCEL logs are deemed more suitable to capture knowledge-intensive processes as they unambiguously link attribute value changes to the correct objects and events with the use of foreign keys.
A snippet of a DOCEL log applied to this running example is shown in Tables \ref{DOCEL Event Table}, \ref{Publisher Table} and \ref{Publication Status Table}. The objects tables of the \texttt{Author} and \texttt{Book} object types are similar to the \texttt{Publisher} object table. Given a DOCEL log of the running example, the aim of this research is to discover:

\begin{itemize}
    \item the DRD graph of Figure \ref{DMNbookrunning} so as to know how the decisions driving the process are structured.
    \item the decision logic driving each subdecision, i.e. discovering the logic of the decisions \texttt{Book's Quality} and \texttt{Publish or not} (see Figure \ref{DMNbookrunning}).
    \item Discovering which activity in the business process alters and introduces the variables needed to make a decision, e.g., activity \texttt{Read manuscript} creates the \texttt{Reviewer Score} and activity \texttt{Decide on publication} alters the variable \texttt{Publication status}.
    \item Discovering to which objects which variables belong to, e.g., \texttt{Total number of published books} belongs to the object type \texttt{Author} whilst the variable \texttt{Review Score} belongs to the object type \texttt{Book}.
\end{itemize}
\begin{table}[]
\caption{Snippet of DOCEL Event Table}
\label{DOCEL Event Table}
\centering
\resizebox{0.8\textwidth}{!}{%
\begin{tabular}{|l|l|ll|l|l|l|}
\hline
\textbf{EID} & \multicolumn{1}{c|}{\textbf{Activity}} & \multicolumn{2}{l|}{\textbf{Timestamp}} & \textbf{Authors} & \textbf{Books} & \textbf{Publishers} \\ \hline
e1 & Find inspiration & \multicolumn{2}{l|}{2022-01-01 10:53:45} & a951 &  &  \\ \hline
e2 & Find inspiration & \multicolumn{2}{l|}{2022-01-01 10:59:42} & a155 &  &  \\ \hline
e3 & Write book & \multicolumn{2}{l|}{2022-01-01 11:00:30} & a951 & b2 &  \\ \hline
e4 & Submit book manuscript & \multicolumn{2}{l|}{2022-01-01 11:08:23} & a951 & b2 & p90 \\ \hline
es & Write book & \multicolumn{2}{l|}{2022-01-01 11:08:32} & a155 & b3 &  \\ \hline
e6 & Read manuscript details & \multicolumn{2}{l|}{2022-01-01 11:17:26} &  & b2 & p90 \\ \hline
e7 & Submit book manuscript & \multicolumn{2}{l|}{2022-01-01 11:18:13} & a155 & b3 & p86 \\ \hline
e8 & Read manuscript details & \multicolumn{2}{l|}{2022-01-01 11:22:06} &  & b3 & p86 \\ \hline
e9 & Read manuscript & \multicolumn{2}{l|}{2022-01-01 11:27:06} &  & b2 & p90 \\ \hline
e10 & Read manuscript & \multicolumn{2}{l|}{2022-01-01 11:31:22} &  & b3 & p86 \\ \hline
e11 & Determine book quality & \multicolumn{2}{l|}{2022-01-01 11:36:09} & a951 & b2 & p90 \\ \hline
e12 & Determine book quality & \multicolumn{2}{l|}{2022-01-01 11:36:51} & a155 & b3 & p86 \\ \hline
e13 & Decide on publication & \multicolumn{2}{l|}{2022-01-01 11:38:02} &  & b3 & p86 \\ \hline
\end{tabular}%
}
\end{table}
\begin{table}
\begin{minipage}{0.49\linewidth} 
\centering
\tiny 
\caption{Snippet of DOCEL Publisher Object Table}
\label{Publisher Table}
\begin{tabular}{|l|l|l|}

\hline
\textbf{PID} &
  \textbf{Name} &
  \textbf{\begin{tabular}[c]{@{}l@{}}Publisher \\ Specialty Genre\end{tabular}} \\ \hline
p44 &
  \begin{tabular}[c]{@{}l@{}}Source Industries\\ Universal Publishing\end{tabular} &
  Romance \\ \hline
p86 & \begin{tabular}[c]{@{}l@{}}Telecom \\ Electronics Publishing\end{tabular} & Romance \\ \hline
p90 & \begin{tabular}[c]{@{}l@{}}Studio West\\ Software Publishing\end{tabular} & Fantasy \\ \hline
\end{tabular}
\end{minipage}%
\hfill
\begin{minipage}{0.49\linewidth} 
\centering
\tiny
\caption{Snippet DOCEL Publication Status Table}
\label{Publication Status Table}
\begin{tabular}{|l|l|l|l|}
\hline
\textbf{PSID} & \textbf{Publication Status} & \textbf{EID} & \textbf{BID} \\ \hline
ps1 & Pending & e19 & b1 \\ \hline
ps2 & No & e23 & b1 \\ \hline
ps3 & Pending & e4 & b2 \\ \hline
ps4 & Yes & e14 & b2 \\ \hline
ps5 & Pending & e7 & b3 \\ \hline
ps6 & Revise & e13 & b3 \\ \hline
\end{tabular}
\end{minipage}
\end{table}

\subsection{Problem Statement}
With a knowledge-intensive process log, discovering decisions can help decision-makers understand how these are executed in the business process.
This way processes and decisions can be changed and analyzed independently of one another  \cite{holmberg2014purity}. 
Additionally, linking decisions to the control flow provides a holistic view of the process, showing when decision-related activities occur. 

Just as it is valuable to discover multiple object types and variables in a process  \cite{van2001proclets}, it is also important to identify the object types and variables involved in decision-making process.
However, this information is not stored in single-case process event logs such as XES logs and can therefore not be discovered in such logs. Secondly, the only other holistic decision discovery algorithm  \cite{de2019holistic} is not applicable to an object-centric process log as it was designed with a single process point of view in mind.
Lastly, currently the main object-centric process discovery algorithms focus on the control flow without decision information  \cite{van2020discovering,berti2022oc}. 
In short, given an object-centric event log, the goal of this research is to:

\begin{enumerate}
    \item \label{RQ1} Discover which activities introduce decision-relevant variables and executing decisions.
    \item \label{RQ2} Discover to which object types the relevant input variables belong to. As such, the object types are also connected to the activities in which they participate.
    \item \label{RQ3} Discover a DMN model for the discovered decisions with both the decision structure and decision logic.
\end{enumerate}

\section{Preliminaries} \label{Preliminaries}

This section first introduces the terminology relevant for the proposed algorithm, next DMN and its terminology are explained.

\subsection{Object-Centric Processes}

\begin{definition} (Data-aware Object-Centric Event Log). A data-aware object-centric event log is a tuple
\sloppy $L=(E, OT, O, A, AT, AV, \pi_{ot},\pi^o_{at},\pi_{id},\pi_a, \pi^e_{at}, \pi_t, \pi_o,\pi_{av},\pi^{e,o}_{at})$:
    \begin{itemize}
        \item $E$ is the set of events, $OT$ is the set of object types (e.g., \{\texttt{Author, Publisher,Book}\}), $O$ is the set of objects, $A$ is the set of activities, $AT$ is the set of attributes, $AV$ is the set of attribute values.
        \item  Each $o_i\in O$ is a tuple $o_i=(oid,ot_i,AT_i)$ where: 
            \begin{itemize}
                \item $oid$ is a unique object identifier (e.g., \texttt{\{p44,p90,a951,b2\}}).
                \item $\pi_{ot}(o_i)=ot_i$ assigns the object to an object type (e.g., $\pi_{ot}($\texttt{p44}$)=$\texttt{Publisher}).
                \item $\pi^o_{at}(o_i)=AT_i$ assigns a subset of attributes to an object with  $AT_i\subseteq AT$ (e.g., $\pi^o_{at}($\texttt{p44}$)=$ \{\texttt{Name, Publisher Specialty Genre}\}).
            \end{itemize} 
        \item Each attribute $at_i\in AT$ is assigned a value $\pi_{av}(at_i)=av\in AV$. This allows us to obtain the values of attributes of events and objects:
        \begin{itemize}
            \item $\pi_{av}(\pi^e_{at}(e_i))= av$ for static event attributes.
            \item $\pi_{av}(\pi^o_{at}(o_i))= av$ for static object attributes (e.g., $\pi_{av}(\pi^o_{at}($\texttt{p44}$))=$ \{{\texttt{Source Industries Universal Publishing, Romance}\}}).
            \item $\pi_{av}(\pi^{e,o}_{at}(e_i,o_i))= av$ for dynamic attributes (e.g., $\pi_{av}(\pi^{e,o}_{at}(e_{19},\texttt{b1}))=$ \texttt{Pending}) which are the sets of attributes that have been changed by an event belonging to an object.
        \end{itemize}
        \item  Each event $e_i\in E$ is a tuple $e_i=(id_i,a_i,at_i,t_i,o)$ where $a_i\in A$ is an activity, $at_i\subseteq AT$ a set of attributes, $t_i \in \mathbb{N}$ a timestamp, $o\subseteq O$ a set of objects. We write:
        \begin{itemize}
            \item $\pi_{id}(e_i)=id_i$ assigns a unique ID to each event. (e.g. $\pi_{id}(e_1)=e_1$).
            \item $\pi^e_{at}(e_i)=at_i$ assigns attributes to events (e.g., $\pi^e_{at}(e_1)=$ \texttt{resource}).
            \item $\pi_a(e_i)=a_i$ assigns an activity to each event (e.g., $\pi_a(e_1)=$ \texttt{Find inspiration}).
            \item $\pi_t(e_i)=t_i$ assigns a timestamp to each event (e.g., $\pi_t(e_1)= 2022{-}01{-}01$  $10{:}53{:}45$).
            \item $\pi_o(e_i)=o$ assigns a set of objects to each event (e.g., $\pi_o(e_1)=$\texttt{a951}). 
        \end{itemize} 
        
        \end{itemize}
\end{definition}

\subsection{DMN}

DMN and its corresponding models were introduced in 2015 by the Object Management Group (OMG) as a Decision Model and Notation standard to model, communicate and execute operational decisions  \cite{DMN2015}. Figure \ref{DMNbookrunning} shows a DMN model dealing with whether or not to publish a book. Every DMN model consists of two parts, one part visualizes the structure of a decision (left side of Figure \ref{DMNbookrunning}), the other part represents the decision logic (right side of Figure \ref{DMNbookrunning}). 

To represent the decision structure, DMN uses Decision Requirement Diagrams (DRD) where a decision such as \texttt{Publish or not} or \texttt{Book's Quality} is represented with rectangles whilst the inputs of decisions such as \texttt{Author's number of published books} are represented with rounded rectangles. Inputs and decisions are connected with solid arrows which are also called information requirements. Finally, it is possible for the output of a decision to be used as an input for another decision, e.g., the output of \texttt{Book's Quality} is used as input for decision \texttt{Publish or not}.
The following terminology is introduced:

\begin{definition} Below we provide the relevant definitions for a DMN model.
    \begin{itemize}
        \item \sloppy $D$ is the universe of decisions and $d$ is a tuple defined as $d=(I_d,O_d,a_d,\pi_{adec},\pi_{out})$ with:
        \begin{itemize}
            \item $I_d \subseteq AT$, the input variables (e.g. \texttt{Reviewer Score, Author's number of published books} are inputs of \texttt{Book's Quality}).
            \item $O_d \subseteq AT$, the output variables (e.g., \texttt{Book's Quality} is the output of the decision \texttt{Book's Quality})
            \item $a_d \in A$, the activity executing the decision (e.g.,  \texttt{Determine book quality} executes the decision \texttt{Book's Quality}).
            \item $\pi_{adec}: A_d \rightarrow D$ is the function associating an activity to a decision (e.g. $\pi_{adec}: \texttt{Determine book quality} \rightarrow \texttt{Book's Quality}$). Every decision needs to be linked to maximum one activity but not each activity needs to be linked to a decision.
            \item $\pi_{out}:  2^{AT} \rightarrow AT$ is a function producing an output variable taking as input the input variables (e.g. $\pi_{out}:$ \{\texttt{Reviewer Score, Author's number of published books}\} $\rightarrow$ \texttt{Book's Quality} with a possible rule:  if \texttt{Total number of published books} $<$ 5 and \texttt{Reviewer Score} in [0..5] then \texttt{Book's Quality} = Bad.)
        \end{itemize}
         \item A DRD is defined as a tuple with $DRD=(D_{drd},I_{drd},R_{drd})$ with: $D_{drd} \in D$ all the decisions in the DRD, $I_{drd} \in AT$ all the input variables of $D_{drd}$, $R_{drd}: (I_{drd} \times D_{drd}) \cup (D_{drd} \times D_{drd}$) the connection between the inputs or intermediary decisions with the decisions to one another.
    \end{itemize}
    
\end{definition}

The decision logic can be represented with decision tables in DMN as they allow for consistency, correctness and completeness of decision rules  \cite{vanthienen1998illustration}. Each rule in a decision table is read as an IF-THEN statement, e.g., in the decision table \texttt{Publish or not}, the third rule must be read as: If \texttt{Book's Quality} is average and if \texttt{Complying with Publisher requirements} is true, then the \texttt{Publication Decision} is revise. A complete overview of DMN specifications can be found in  \cite{DMN2015}.

\section{IODDA Algorithm} \label{P-MinD Algorithm}

This section introduces the Integrated Object-centric and Decision Discovery Algorithm (IODDA), which aims to uncover decisions in an event log. IODDA is a three-step algorithm following the general idea of any data analysis project namely: pre-processing, analysis, and post-processing. The general steps of IODDA are:
\begin{itemize}
    \item Step 1 (Pre-processing): For each activity discover a set of potential input variables ($I_d$) and potential output variables ($O_d$) belonging to the relevant object types.
    This step is important as it delineates the relevant dimensions in which IODDA will have to perform a more in-depth analysis.
    The pre-processing step is broadly responsible for solving research goal \ref{RQ1}.
    \item Step 2 (Analysis): Determine with a correlation metric value $C$ which variables are used as inputs by activities that change the values of output variables.
    Next, for each potential input-output combination, a prediction model is built.
    In this step, IODDA builds a proxy model $L$ for $\pi_{out}: AT \rightarrow AT$ which is supported with a sufficiently high value of correlation metric $C$.
    This second step completely solves research goal \ref{RQ2} and for research goal \ref{RQ3}, it builds candidate DMN models with their structure and logic.
    \item Step 3 (Post-processing): 
    Since the previous step might discover decision models which overlap with one another because it analyzes different objects without always taking into account what is already discovered, this step ensures that the information is consolidated in fewer but completer decision models.
    The last step finalizes research goal \ref{RQ3} by creating final DMN models.
\end{itemize}

Before IODDA is explained in further detail, a key underpinning concept of this algorithm called a shift is introduced.
\begin{definition}
    A shift is a function $\sigma^{at,a}_o= AT \times A \times O \to E$ capturing all the events of an activity $a$ changing an attribute $at$ of object $o$. \\
    e.g., $\sigma^{Publication Status, Decide on publication}_{b3}= e_{13}$.
\end{definition}

To refer to an individual shift function, the rest of the paper also refers to them as Attribute-Activity-Object\_Type shift (ATAOTSs), e.g., (Publication Status, Decide on publication, Book,2)-shift refers to the idea that the activity \texttt{Decide on publication} modified the attribute \texttt{Publication Status} of object type \texttt{Book} for the second time.

Similarly $\sigma_o= O \to E$ captures all events in which object $o$ is present.
This shift concept is created the first time the IODDA algorithm goes through a DOCEL log.
This  allows for an efficient analysis in the remaining of the algorithm as otherwise IODDA would have to go through the entire object-centric log multiple times in search of the same information.
Finally, within the context of IODDA, a discovered DRD has to be defined as follows:
\begin{definition}
    $\textit DRD_{disc} = (D_{drd},I_{drd},R,L,C,T_{{at}_o})$ where $D_{drd}$ and $I_{drd}$ denote ATAOTS with n shifts. $R$ represents input-output edges, $L$ is the predictive model, $C$ signifies input-output correlation, and $T_{at_o}$ represents relevant attribute-activity-object shifts.
\end{definition}

\subsection{Step 1: Discovering potential input and output variables}

In the initial phase of IODDA, it identifies potential input and output variables for each activity. These input variables are then placed into potential decision models for further analysis if they have a potential impact on the output variable.

In more detail, the algorithm is initialized with an empty set $DRD$ and $\sigma^{at,a}_o=\emptyset$ an empty set of attribute-activity-object shifts. 

Next, in lines 4-12, all objects of all object types are analyzed with the assumption that the first event in which an object appears also creates its static attributes $AT^{static}_o$ (lines 7-9). For the dynamic attributes $AT^{dynamic}_o$, the algorithm creates a shift function, $\sigma^{at_{dynamic},a}_o$, returning the events for each (activity, attribute)-combination if the currently analyzed event also changed the value of the dynamic attribute (lines 10-12).

Once all the ATAOTSs have been identified and $\sigma^{at,a}_o$ is complete, lines 13-20 identify for each activity a set of potential input and output variables of all object types of that activity.
A variable is considered a potential input variable of an activity if it has more than one distinct value.
Next, to qualify as an output variable two conditions are checked:
\begin{enumerate}
    \item the variable has more than one distinct value.
    \item the variable must be shifted a minimum number of times relative to the number of objects of a specific type across all activity executions. This minimum shift threshold is determined by the parameter called $\underline{minShift}$. For example, if 10 events modify a variable for 10 different objects of type \texttt{Book} during an activity, and $\underline{minShift}=0.2$ then if there are a total of 20 books, it is considered significant because more than 20\% of all \texttt{Book} objects were modified. However, if there are 200 books, it's not significant as less than 20\% of the books were affected, and IODDA does not identify it as a potential output variable for the activity. 
\end{enumerate}

This last criterion ensures that only attributes with sufficient value changes ($> \underline{minShift}$) are considered as output variables for an activity.  A further discussion on hyperparameter sensitivity can be found in Section \ref{Discussion}. 

Finally, lines 21-25 analyze all discovered shifts and store these together in $T_{at_o}$ consisting of (activity, attribute, number of shifts).
Each ATAOTS is then proposed as a top node of a new DRD.  Subsequently, the algorithm proceeds with the recursive \textit{Find all input variables} algorithm to identify all input variables associated with the top decision. Finally, any duplicate DRDs are eliminated, and only the unique DRDs are returned as the output of the algorithm. The object-centric aspect in this first part is mainly manifested by the extra loops the algorithm performs for each object type to discover the shifts. Secondly, all the shifts are enriched with their object type information which is necessary information later on to correctly visualize the DRDs.

\begin{algorithm}
\caption{Discovering potential input and output variables}\label{algXESOCEL}
\begin{algorithmic}[1]
\Procedure{MINE\_DMN\_MOD}{$\mathcal{L}$}
\State \textit{DRD} $\leftarrow \emptyset$ \Comment{\textit{DRD} used as global variable throughout algorithms}
\State $\sigma^{at,a}_o \leftarrow \langle \rangle$, 
        $\forall(at,a,o,e) \in (AT \times A \times O) \to E$
        
\For{$ot \in OT_\mathcal{L}$} \Comment{All object types in the log}
    \For{$o \in ot$}    
        \For{$e_t \in \sigma_o$} \Comment{All events object o is involved in}
            \If{$e_{t-1}= \emptyset$} 
                \For{$at \in AT^{static}_o$} 
                    \State $\sigma^{at_{static}}_o \leftarrow e_t$ \Comment{add all static attributes of o to shift} 
                 
                 \EndFor
            \EndIf
            \For{$at \in AT^{dynamic}_o$} \Comment{All dynamic variables of object}
                \If{$e_{at} = e_t$ \AND {$o_{at}=o$}} 
                    \State $\sigma^{at_{dynamic},a}_o \leftarrow e_t$  \Comment{add the dynamic attributes of o to shift} 
                \EndIf
            \EndFor   
        \EndFor
    \EndFor    
\EndFor

\For{$ot \in OT_\mathcal{L}$}
    \For{$o \in ot$} 
        \For{$at \in AT_o$}
            \For{$a \in A$} 
                \If{$| \{val_{at}(\sigma) |\sigma \in \Sigma^{at,a}_o | > 1$} \Comment{The variable is discriminative}
                    \If{$|\sigma^{at,a}_o| > \underline{minShift} \cdot |ot|$} \Comment{enough shifts occurred for ATAOTS}
                    \State $O_a \leftarrow O_a \cup at$
                    \EndIf
                \State $I_a \leftarrow I_a \cup at$    
                \EndIf
            \EndFor
        \EndFor
    \EndFor
\EndFor

\For{$ot \in OT_\mathcal{L}$}
     \Comment{Find all inputs set by other prior activities}
    \For{$(at,a,ot,n) \in \{(at,a,ot,n) \in AT \times A \times OT \times \mathbb{N} | \sigma^{at,a}_o \neq \emptyset \wedge n \in [1, \underline{maxShift_o}] \}$} 
        \State $T_{at_o} \leftarrow \{\sigma | \sigma^{at,a}_{\sigma_o,n} \neq \emptyset \} $ \Comment{All object traces with at least n shifts}
        \State $\textit{DRD} \leftarrow \textit{drd} = (D \cup d = (at,a,ot,n),I_{drd},R,L,C,T_{v_o})$
        \State Find input variables$(at,a,ot,n,drd)$
    \EndFor
\EndFor
\State $\textit{DRD} \leftarrow$ eliminate duplicates$(\textit{DRD})$
\State \Return DRD
\EndProcedure
\end{algorithmic}
\end{algorithm}

\subsection{Step 2: Building decision models}

\begin{algorithm}
\caption{Building decision models}\label{alginputs}
\begin{algorithmic}[1]
\Procedure{Find Input variables}{$d=(at,a,ot_i,n), drd=(D,R,L,C,T_{{at}_o})$} 

\State $M,D_2 \leftarrow \emptyset$ 

\State $\textit{ATAOTS} \leftarrow \{(at_2,a_2,ot_2) \in AT \times A\times OT) | at_2 \in I_d \wedge at_2 \in \{at | at \in O_{a_2}\}\} $

\For {$(at_2,a_2, ot_2) \in ATAOTS$} 
    \For{$d_2=(at_2,a_2,ot_2,n_2) \in \{(at_2,a_2,ot_2,n_2) \in AT \times A \times OT \times \mathbb{N} |  \sigma^{at_2,a_2}_{o_2} \neq \emptyset \wedge n_2 \in [1,\underline{maxShift}]\}$}

        \State $val_{at}, val_{at_2}, T \leftarrow \emptyset $ 

        \For{$\sigma \in T_{{at}_o}$} \Comment{Loop over all object traces of the top activity}
            \If{$o_2 \in a \wedge at_2 \in AT_{o_2} \wedge \sigma^{at_2,a_2}_{o_2,n_2} < \sigma^{at,a}_{o,n} $}
                \State  $val_{at} \leftarrow val_{at} \cup  val_{at}(\sigma^{at,a}_{o,n}), val_{{at}_2} \leftarrow val_{{at}_2} \cup  val_{{at}_2}(\sigma^{at_2,a_2}_{o_2,n_2}),
                T \leftarrow T \cup \sigma$
            \EndIf
            \If{$|T| > |OT_i| \cdot \underline{minTraceprop} \wedge corr(val_{at}, val_{{at}_2}) > \underline{minCorr}$} 
            \State $D2 \leftarrow d_2$
            \State $M \leftarrow (d_2,T)$
            \State $C(d,d_2) \leftarrow cor(val_{at},val_{{at}_2})$
            \EndIf
        \EndFor
    \EndFor

\EndFor
\State Sort(M)

\For{$(D_2,T) \in$ Find input models(M)}
    \State $L(D_2,at) \leftarrow $ Build predictive model($D_2,at,T$) 
    \If{$L(D_2,at) > \underline{minSupport}$}
        \For{$d_2 \in D_2$}
            \State $D \leftarrow d_2$
            \State $R \leftarrow (d_2,d)$
            \If{$T=T_{{at}_o}$}
                \State Find input variables($o_{d_2},a_{d_2},OT_2,n_{d_2},drd$)
            \Else  \Comment{Other object trace cluster requires different model to be built}
            \State re-estimate $C$ for $T$ and adjust $R$ accordingly to obtain $C_n$, $R_n$
            \State $DRD \leftarrow drd_n= (D,R_n,L,C_n,T)$
            \State Find input variables$(o_{d_2},a_{d_2},OT_2,n_{d_2},drd_n)$

            \EndIf
        \EndFor
    \EndIf
\EndFor
\State \Return $(D,R)$
\EndProcedure
\end{algorithmic}
\end{algorithm}

The second part develops candidate models by recursively discovering which candidate input variables are sufficiently correlated with their candidate output variables across all present object types.
In this part, it is important to correctly link the involved objects of different object traces to one another otherwise the algorithm might incorrectly link attribute values to unrelated objects and as such discover wrong object-object relationship maybe leading to wrong decision models later on. 

In lines 2-3, the algorithm initializes three key components: the empty set of decision models $M$, the empty set of potential input variables for the top node namely $D_2$, and a set of relevant ATAOTSs.
An ATAOTS is considered relevant if the associated variable serves as an input to the top node activity $a$ and as an output of the activity $a_2$ of the potential input variable.

Lines 4-9 identify which object traces experienced shifts that subsequently influence the variables of the top decision.
By employing backward reasoning on the top activity, the algorithm checks for each involved object in the top decision activity whether any object input variable was created or altered before the top activity alters the output variable.
If such a case is found, the set of values are respectively stored in $val_{at}$ and $val_{{at}_2}$ and the relevant object traces  are stored in \textit{T}.
The parameter $\underline{maxShift}$ allows to determine how many times a potential input variable may have been changed in the past to be considered relevant for further analysis. This allows for a more flexible analysis on various process logs.

Proceeding to lines 10-14, the algorithm first checks if the number of object traces in $T$ for which the potential input variable was changed or created before the output variable exceeds a predetermined value (\textit{\underline{minTraceprop}}) relative to the total number of objects of the corresponding object type to which the output variable belongs to. 
This mirrors the concept of \textit{\underline{minShift}} in identifying significant input variables compared to the object population.
Next, it checks if the correlation between the top decision (output variable) and the potential input variable exceeds a set threshold (\textit{\underline{minCorr}}). If both criteria are met, the potential input variable, traces, and correlation are stored.
Next, the proposed models are ranked based on decreasing number of covered object traces.
The algorithm uses mutual regression and classification information as proxies for determining variable correlations  \cite{manning2009introduction}.  However, IODDA can work with any inference technique and thus also with causal machine learning techniques such as uplift trees \cite{rzepakowski2010decision}. IODDA has a plug and play mechanism regarding inference techniques and can serve as a basic structure which can be further used with causal graphs. However, DMN tables do not aim to contain causal rules. Moreover, it is not guaranteed that a certain change can solely be attributed to an individual variable if it is part of a rule with multiple variables. Therefore, we use the correlation metric as proxy to extract decision rules which in turn need to be validated by experts.

Continuing to lines 15-27, the algorithm builds a predictive model \textit{L} using the corresponding potential input variables, the object traces and the top decision.
Random forests are employed in this case due to their robustness, and performance, although alternative predictive models could be utilized such as XGBoost or neural networks.
The predictive model's binary accuracy is evaluated, and if it surpasses a defined threshold \textit{\underline{minSupport}}, the algorithm proceeds to recursively identify further potential input variables of the newly determined input variable $D_2$.

If the object traces of the input decision do not completely overlap with the object traces of the top node, the algorithm recalculates correlations and edges for overlapping traces of both the input and output variables before proceeding with the recursive search for new input variables.
This aspect is especially important in an object-centric context where the total number of objects can vary between object types.
Because of this objects can participate in the decision-making of various other objects belonging to other object types.
This makes that not all object traces are only linked to a single other object trace, it is therefore important to know how and which objects are related to one another otherwise certain variables might receive too much or too little importance because of an incorrect object-object relationship analysis.
Ultimately, the algorithm returns all decisions and their associated edges, providing a comprehensive output for further analysis.

\subsection{Algorithm 3: Finding all overlapping models}

\begin{algorithm}
\caption{Finding all overlapping models}\label{algoverlap}
\begin{algorithmic}[1]
\Procedure{Find input models}{$M$} \Comment{M: set of potential models}

\State $models \leftarrow \emptyset$ \Comment{Generated models}
\State $cov \leftarrow \emptyset$ \Comment{Covered object traces}

\For{$((at,a,OT,n),T) \in M$}
    \For{$(D,T_m) \in models$} \Comment{Check all possible input models}
        \If{$T=T_m$} \Comment{If traces are the same, add to model}
            \State $D \leftarrow \cup (at,a,OT,n)$
        \ElsIf{$T \subset T_m \wedge |T| > OT \cdot \underline{minTraceprop}$}
        \State $models \leftarrow (D \cup (at,a,OT,n),T)$
        \ElsIf{$T \cup T_m \neq \emptyset \wedge \frac{|T \cup T_m|}{|T|} > \underline{minDev} $}
        \State $models \leftarrow (D \cup (at,a,OT,n),T \setminus T_m )$
        \State $cov \leftarrow T$
        \EndIf
    \EndFor
    \If{$T \subsetneq cov \wedge |T| > OT \cdot \underline{minTraceprop}$}
    \State $models \leftarrow ((at,a,n),T) $
    \State $cov \leftarrow T$
    \EndIf
    
\EndFor
\State \Return $models$

\EndProcedure
\end{algorithmic}
\end{algorithm}

IODDA's final part ensures that the discovered models are combined effectively, taking into account the proportional coverage of object traces relative to the total number of relevant object traces and the compatibility of variables.
Again, this is tied to object traces influencing other types and certain variables being vital to object types due to their relative importance rather than sheer quantity.
By examining various overlapping object trace scenarios and applying preset thresholds, the model includes all relevant input variables while preserving predictive power and model coherence.

The algorithm loops over all ATAOTSs that have been identified as potential output variables or top decisions previously.
The initialization of a model occurs when enough object traces are covered proportionally to the total number of that object type as determined by the \underline{\textit{minTraceprop}} parameter.
Each potential model is stored in the \textit{models} set and the object traces covered by the model are added to the \textit{cov} sets, as outlined in lines 13-15.

In lines 5-12, the algorithm evaluates each potential input variable's compatibility with the existing model. Three scenarios are considered. First, if the input variable's covered object traces fully match the current overall model, it is incorporated into the model as an input variable of the top decision, as shown in lines 6-7.
Secondly, if the covered traces of the input variable partially overlap with its belonging output variable but cover sufficient object traces in proportion to the object type determined by \underline{\textit{minTraceprop}}, the input variable is included (lines 8-9).
Lastly, if two sets of object traces do not entirely overlap with the same objects but are similar enough determined by \underline{\textit{minDev}}, meaning they can be traced back to common objects, the input variable is still added to the combined model if it adds enough predictive power (lines 10-12).

\section{Artificial DOCEL log generators} \label{Artificial DOCEL logs}

Public OCEL logs lack descriptions of decision-making processes, and there are currently no publicly available DOCEL logs detailing decisions. To support future research in object-centric process analysis, two knowledge-intensive object-centric DOCEL process log generators were developed. The first process corresponds to the publication process in Section \ref{Problem Statement}, while the second pertains to the order shipping method described in the next section.
Each log generator allows customization of parameters to produce logs with varying behaviors.

\subsection{Shipping Method Example}

\begin{figure*}[]
\centering
\includegraphics[width=1\linewidth]{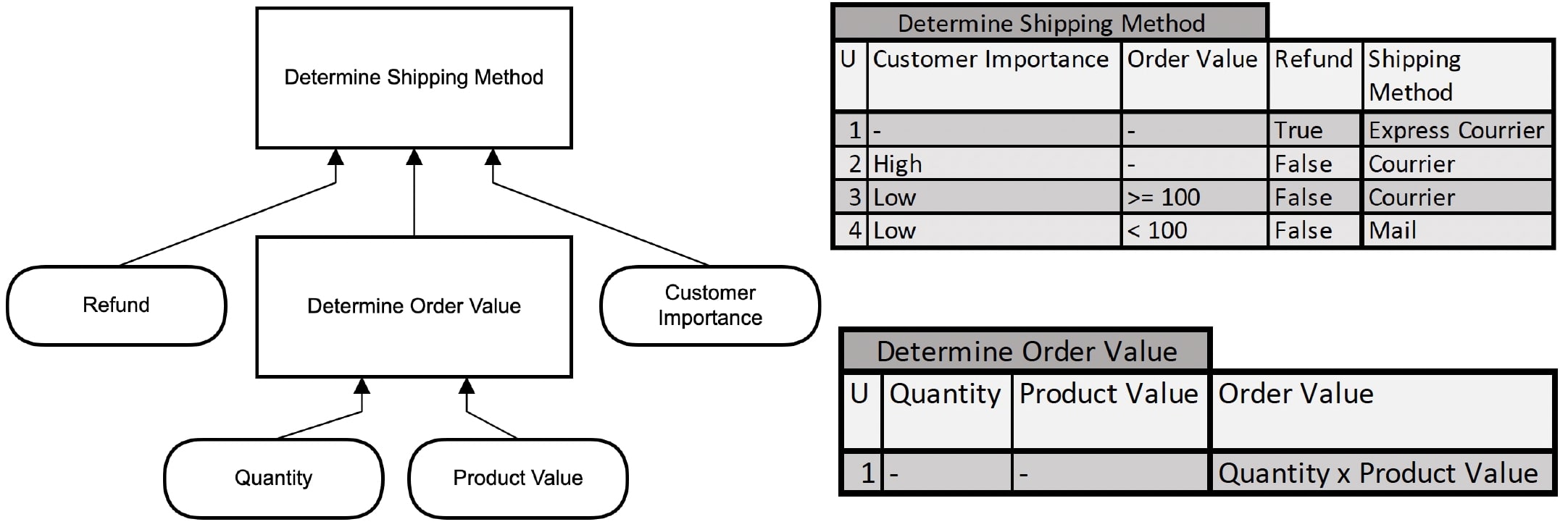}
\caption{Shipping Method DMN model} \label{DMNship}
\end{figure*}

A second problem is introduced dealing with an order-shipping method process inspired by the process described in  \cite{de2019holistic}. The BPMN model 
can be found on the GitHub page\footnote{https://github.com/Goossens496/IODDA.git} 
and the corresponding DMN model is shown in Figure \ref{DMNship}. 

The process contains three object types, i.e., \texttt{customer}, \texttt{order} and \texttt{product type}. Depending on a variety of variables such as the order value, the importance of a customer, and whether or not the customer has requested a refund, a different shipping method is used. A textual description of the process is provided below in italics.

\textit{A customer has been assigned an importance level by the company in the past. The process starts with a customer ordering a certain amount of a product. This is then received by the company which determines the value of the order. Once all the shipping information has been confirmed, the shipping method is determined based on the importance of a customer, the value of the order and on whether the customer has asked for a refund. At the same time, the invoice is send. Once the package has been delivered, the customer is either satisfied or not. If the customer is satisfied, the order is filed and the process ends. Otherwise, a refund is asked and the package is sent back with an express courier. Once the refund has been paid and the package has been successfully delivered, the process is finished.}

The decision logic is shown in Figure \ref{DMNship}. The order value is determined by multiplying the quantity with the product value. The shipping method is determined by the \texttt{Determine Shipping Method} Table. In short, if a customer requests a refund, the shipping method is \texttt{Express Courrier}. If a customer is considered important or the order value $>=100$ then the shipping method is \texttt{Courrier}, otherwise the shipping method is \texttt{Mail}.

\subsection{Artificial Log Generators}

To promote reproducibility and support the research community, the artificial logs, DOCEL log generators, IODDA Python implementation, and complete BPMN model figures are made available and can be found on the GitHub page. The artificial log generators are provided as Jupyter Notebooks with global parameters. Tables \ref{Publication Process Log Generator Parameters} and \ref{Shipping Method Log Generator Parameters} show the editable parameters. The decision logic and thresholds in both log generators can be readily adjusted, allowing customization of event and decision outcome distribution to suit specific research requirements. 
The DOCEL logs themselves are exported to a common spreadsheet format with each sheet representing the events, object or dynamic attributes table.

\begin{table}[]
    \caption{Summary of Functionality in the Publication Process Log Generator}
    \label{Publication Process Log Generator Parameters}
    \centering
    \begin{adjustbox}{width=.8\textwidth}
    \begin{tabular}{|l|p{10cm}|l|p{10cm}|}
        \hline
        \textbf{Tunable parameters} & \textbf{Description of Functionality} \\
        \hline
        Maximum number of Authors & Enables creating new authors with a random number of published books, random genre in which the author is specialized and a random name. \\
        \hline
        Max published books & Maximum number of published books an author can have. \\
        \hline
        Genre list & List of potential genres in which an author or publisher can be specialized in.  \\
        \hline
        Maximum number of Publishers & Enables creating new publishers with a random company name and a specialized genre.  \\
        \hline
        Number of Books & Determines how many books will be written over the course of the simulation.   \\
        \hline
        Start Timeframe & Allows defining the start time for the process. \\
        \hline
        Time interval Between Events & Allows adjusting the time interval between consecutive events.  \\
        \hline
        Probability of Publisher Compliance & Can be adapted to determine the likelihood of a book meeting the initial publisher requirements. The lower the value the more likely a book is compliant. \\
        \hline
        Publication threshold & Determines how many prior publication an author should have in the decision of the quality of submitted book.\\
        \hline
    \end{tabular}
\end{adjustbox}
    
\end{table}
\begin{table}[]
    \caption{Summary of Functionality in the Shipping Method Log Generator}
    \label{Shipping Method Log Generator Parameters}
    \centering
    \begin{adjustbox}{width=.8\textwidth}
    \begin{tabular}{|l|p{10cm}|l|p{10cm}|}
        \hline
        \textbf{Tunable parameters} & \textbf{Description of Functionality}  \\
        \hline
        Product Value & Allows changing the value of the product. \\
        \hline
        Number of Customers & Enables creating new customers with randomly generated names, random belgian bank accounts and randomly assigned importance values (\textit{High or Low}).   \\
        \hline
        Lists of People Executing Activities & Allows adapting the lists of people who execute the company's activities. \\
        \hline
        Start Timeframe & Allows defining the start time for the process.  \\
        \hline
        Time interval Between Events & Allows adjusting the time interval between consecutive events. \\
        \hline
        Order Value Threshold & Can be changed to determine the shipping method based on the order's value. \\
        \hline
        Number of Orders & Can be changed to generate logs with varying numbers of orders. \\
        \hline
        Probability of Refund & Can be adapted to determine the likelihood of a refund, indicating customer satisfaction with the purchase. The higher the value the more satisfied the customer is. \\
        \hline
        Maximum order quantity & Influences how many product units a customer can buy in one order. \\
        \hline
    \end{tabular}
\end{adjustbox}
\end{table}

\section{IODDA Implementation and Demonstration} \label{P-MinD in Action}

The IODDA algorithm is available in Python, along with the log generators on the GitHub page. It has been tested on both artificial DOCEL logs with their specific information regarding number of events, objects, average number of objects per event and attributes described in \autoref{Event log information}.  Given that the algorithm is demonstrated using artificial logs, the ground truth is the logic driving the simulation in the artificial log generators which is defined by the user in this case. However, this does not have to be the case in a real-world application where the logic is simply part of the object-centric log. 
The main challenge of these logs, besides discovering the decisions, is to also correctly uncover the interactions among objects within and across events.

Given that in this study decision trees are built based on the event log these can be considered to be complete and unique for the analyzed objects and valid for the analyzed event log.

The discovered DRDs and decision trees are reported in the following subsections. 

\begin{table}[h]
\centering
\caption{Event log information}
\label{Event log information}
\resizebox{0.6\textwidth}{!}{%
\begin{tabular}{|l|l|l|}
\hline
                      & \textbf{Publication Process} & \textbf{Shipping Method Process} \\ \hline
\textbf{\#Events}     & 800                          & 1596                             \\ \hline
\textbf{\#Object Types} & 3                          & 3                                \\ \hline
\textbf{\#Objects}    & 299                          & 151                              \\ \hline
\textbf{AVG(\#Objects/\#Events)}    & 2.25                        & 1.86                             \\ \hline
\textbf{\#Attributes} & 11                           & 9                                \\ \hline

\end{tabular}
}
\end{table}

The Python implementation's generated DRDs include object-centric and control flow details. It specifies the object type associated with the decision's variable and the influencing activity.
Next, for each relationship between decision elements it is indicated for how many objects this relationship has been observed. Note that the variable names are indicated with \texttt{Name\_shift-x}. For example \texttt{Name\_shift-1}, indicates that IODDA is reporting the results for the first shift of that variable caused by an activity.
For visualizing decision logic, the algorithm employs decision trees because they are widely used and there are more models available for training them. Before constructing the decision tree, the accuracy of the discovered random forest model must surpass a specific threshold. Decision trees can also be converted into decision tables later on \cite{strecht2015survey}. 

\subsection{Book Problem}

\begin{figure*}[]
\centering
\includegraphics[width=\linewidth]{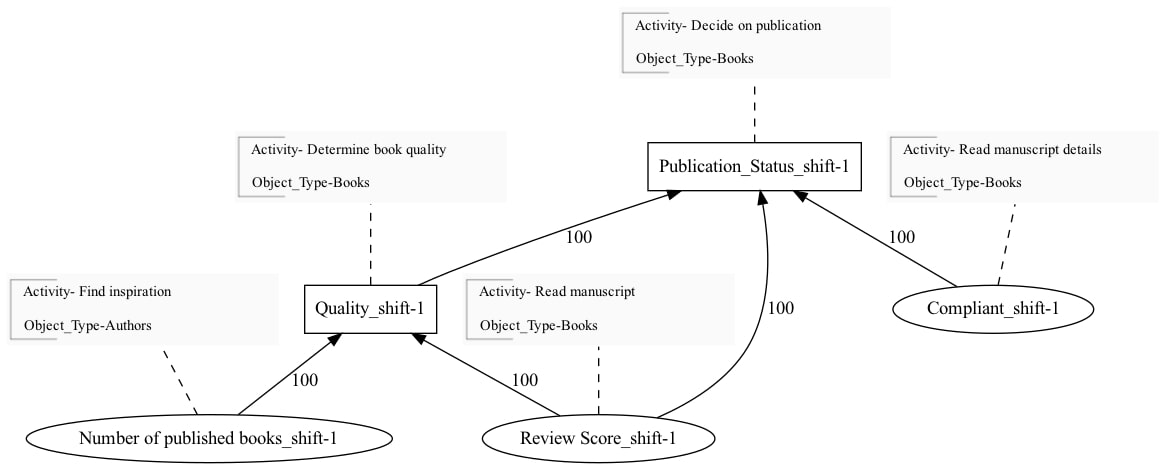}
\caption{Discovered DRD for the book publication problem} \label{PMIND Book DRD}
\end{figure*}

Figure \ref{PMIND Book DRD} illustrates the DRD discovered by IODDA for the publication process, revealing all decisions and relevant information needed to determine the final publication outcome.
The algorithm also correctly discovers that the variable \texttt{Review Score\_shift-1} is created by the \texttt{Read manuscript} activity and belongs to the \texttt{Books} object type.
It also recognizes that, although \texttt{Quality} belongs to \texttt{Books}, the \texttt{Number of published books} of the \texttt{Authors} object type is also relevant in determining \texttt{Quality}.
The numbers above the arrows indicate that the observed relations between the variables and the decisions are valid for all 100 authors and 100 books.
The discovered DRD aligns with the ground truth of Figure \ref{DMNbookrunning}.

\begin{figure*}[]
\centering
\includegraphics[width=1\linewidth]{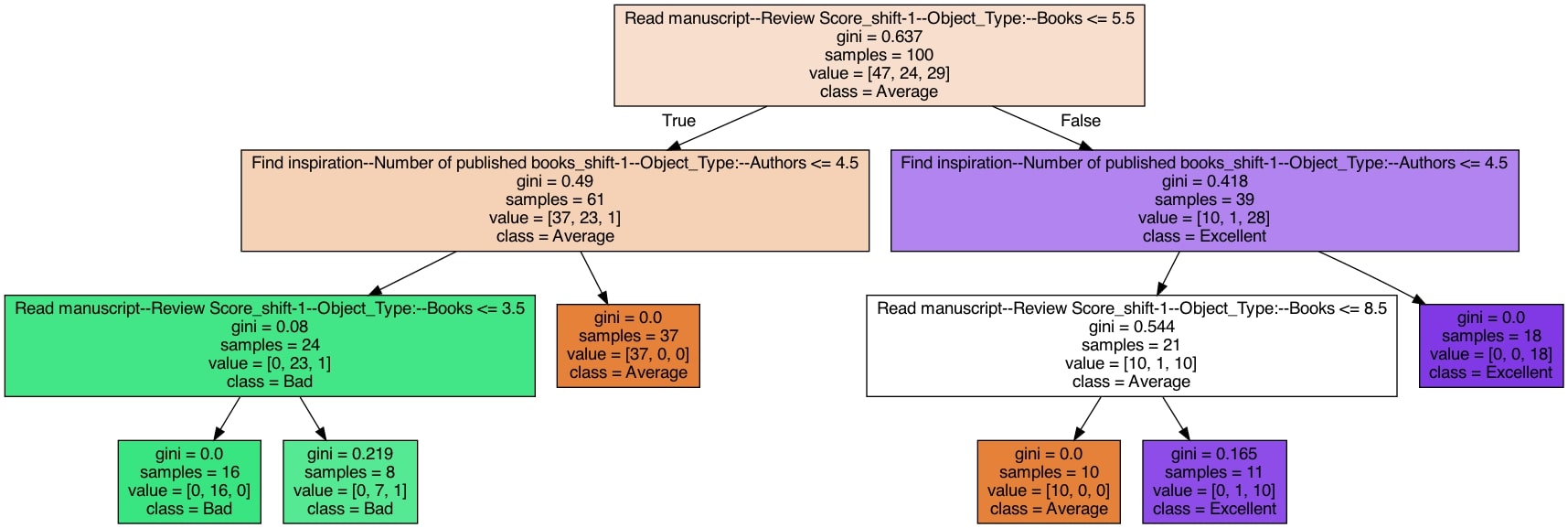}
\caption{Discovered decision tree for the quality decision} \label{PMIND Quality Tree}
\end{figure*}

\begin{figure*}[]
\centering
\includegraphics[width=1\linewidth]{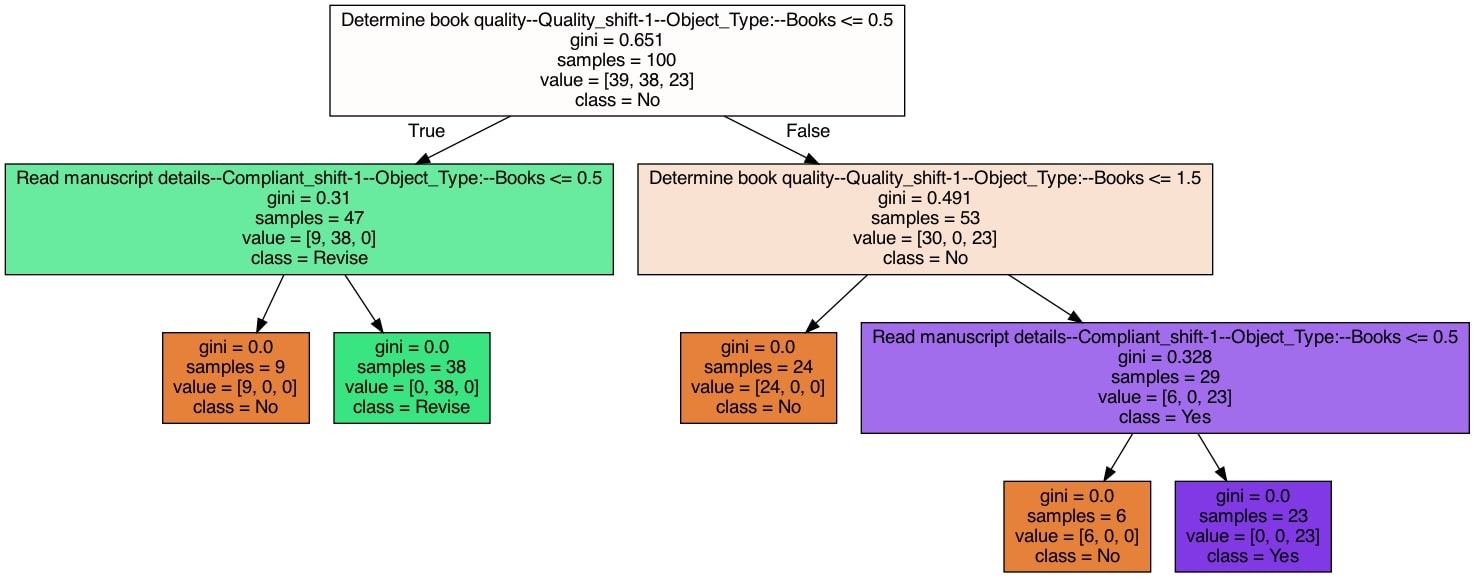}
\caption{Discovered decision tree for the book publication decision}\label{PMIND Publish Tree} 
\end{figure*}

Figures \ref{PMIND Quality Tree} and \ref{PMIND Publish Tree} show the discovered decision tree for the \texttt{Quality} and \texttt{Publication} decisions. 
The decision tree for book quality (Figure \ref{PMIND Quality Tree}) accurately replicates the ground truth of Figure \ref{DMNbookrunning}. 
Notably, the decision tree suggests that \texttt{review score} holds greater importance in determining book quality after considering the \texttt{number of published books by an author}.
For Figure \ref{PMIND Publish Tree}, it is important to know that the algorithm one-hot encoded the \texttt{Quality} variable as follows: 0=Average, 1=Bad, 2=Excellent. The \texttt{Complaint} variable retained its binary values meaning that 0=False and 1=True. 
The decision tree for book publication aligns with the ground truth in Figure \ref{DMNbookrunning}.


\subsection{Shipping Method process}

\begin{figure*}[]
\centering
\includegraphics[width=\linewidth]{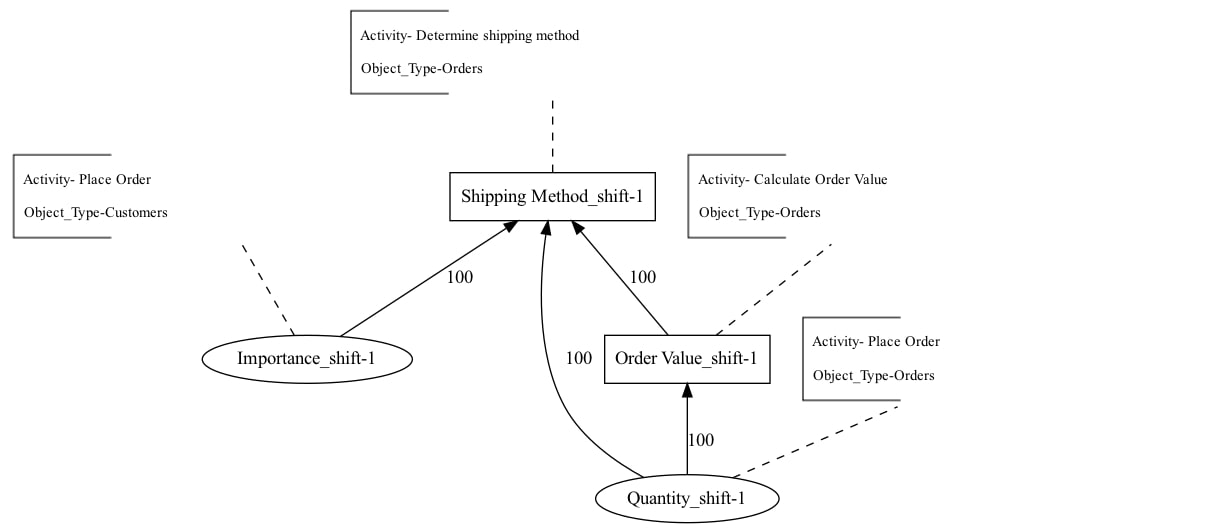}
\caption{Discovered DRD for the shipping method problem} \label{PMIND Ship DRD}
\end{figure*}

The discovered DRD for the shipping method process is shown in Fig \ref{PMIND Ship DRD}.
It accurately identifies that the \texttt{Importance} variable belongs to the \texttt{Customers} object type, while the remaining variables are from the \texttt{Orders} object type.
It correctly identifies that \texttt{Order Value} is solely determined by \texttt{Quantity}  due to the log containing a single product type, therefore \texttt{Product Value} can not influence \texttt{Order Value} since \texttt{Product Value} is a static attribute which does not change values within the log.

\begin{figure*}[]
\centering
\includegraphics[width=.8\linewidth]{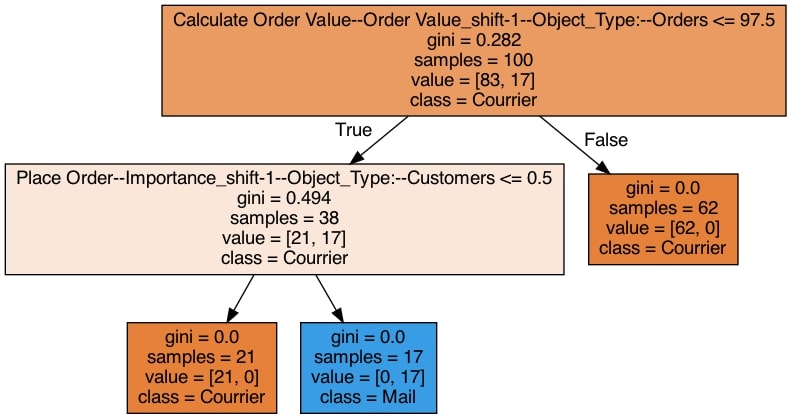}
\caption{Discovered decision tree for the shipping method problem} \label{PMIND Ship decision tree}
\end{figure*}
Figure \ref{PMIND Ship decision tree} visualizes the decision tree for the first shift of the shipping method in the process. Note that the algorithm binarized the \texttt{Importance} variable as follows: 0=High and 1=Low. The algorithm correctly discovers that the \texttt{Order Value} threshold for an order to be send by courrier equals \texttt{100} approximately. It also correctly discovers that when \texttt{Order Value} is below that threshold and that \texttt{Importance} equals low, the shipping method equals \texttt{mail}.

\section{Discussion} \label{Discussion}

The algorithm's DRDs and decision trees reflect the decisions within a log, which can be challenging to discover when the ground truth is absent. 
The discovery of correlations plays a significant role in this process.
For instance, in Figure \ref{PMIND Book DRD} the Review\_Score\_shift-1 and Quality\_shift-1 are connected to the top decision based on their significant correlation, even though the ground truth did not make this connection.
Next, the algorithm is also sensitive to hyperparameter changes, including \underline{\textit{minShift}},\underline{\textit{minTraceprop}},  \underline{\textit{minCorr}}, \underline{\textit{minDev}} and \underline{\textit{minSupport}}.
Adjusting these parameters can greatly affect which object traces and variables are considered noteworthy for further examination.
For instance, adjusting the \underline{\textit{minShift}} to higher values can be interesting for long processes where attributes are changed multiple times over the course of the process.
In case there is a suspicion that a minority of an object type follow a specific pattern of decisions, setting the \underline{\textit{minTraceprop}} to a lower value might help uncover these patterns as it allows for variable changes of less objects to also be analyzed.
Lower values for the \underline{\textit{minCorr}} parameters allow to further analyze more variables even if they have a low correlation value which is relevant in case there are complex decision rules present which are spread across multiple variables since in that case not all variables necessarily have a strong correlation even though they also influence the decision.
Adjusting the \underline{\textit{minDev}} to lower values will allow IODDA to link more loosely connected objects to one another.
Lastly, the \underline{\textit{minSupport}} parameter is important to determine whether the predictive power of the model is strong enough to include the variable as an actual input variable.
In short, lower parameter values result in the discovery of relationships that are present in a smaller amount of object traces, however finetuning the combination of these parameters can greatly influence what is discovered.

\begin{figure*}[]
\centering
\includegraphics[width=\linewidth]{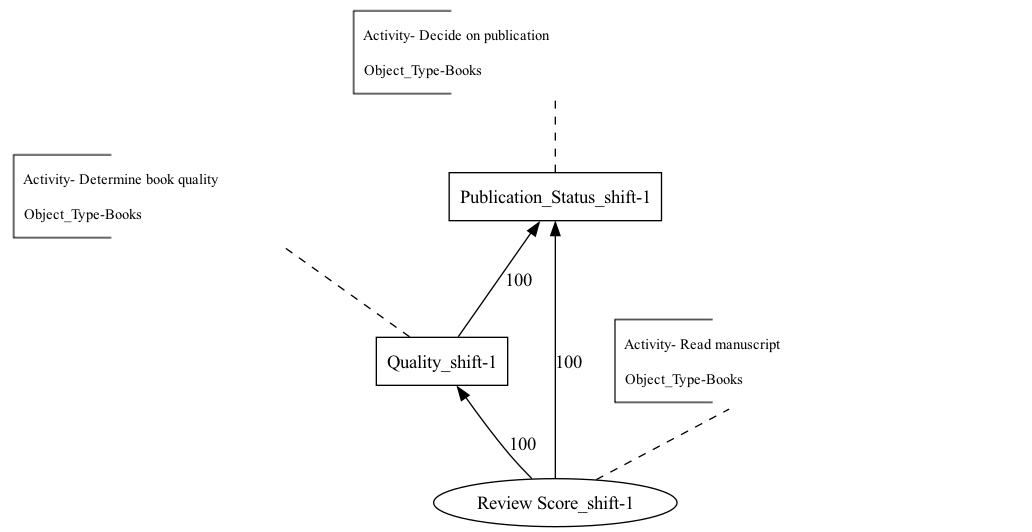}
\caption{Discovered DRD for the publication problem with all hyperparameters set to 0.3} \label{IODDA DRD BOOK 2}
\end{figure*}

We will illustrate the behaviour of IODDA for varying hyperparameter values with the use of the publication problem. Figure \ref{IODDA DRD BOOK 2} shows the discovered DRD in case all the hyperparameters are set to 0.3. These values generate a different DRD with less information which can be used to illustrate the effect of changing an individual hyperparameter value given that the rest remains set at 0.3.

\begin{itemize}
    \item \underline{\textit{minShift}}: 
    will make that an attribute is only considered for further analysis if it has undergone enough shifts relatively to the total number of objects of that object type. Increasing the value of \underline{\textit{minShift}} does not change the discovered DRD since these relationships are valid for all books. Decreasing the value \underline{\textit{minShift}} will also not change the discovered DRD, but under the hood it will consider more object traces to be analyzed given that fewer shifts need to have occurred before being considered for further analysis.
    \item \underline{\textit{minTraceprop}}: Is similar to \underline{\textit{minShift}} but is used to select variables for the decision models. Increasing or decreasing the value does not impact the discovered DRD given that the object traces do not meet the requirements of the other hyperparameters. Note that this hyperparameter is closely linked to the \underline{\textit{minShift}} hyperparameters. Basically, these two hyperparameters should be considered as pairs that need to move together.
    \item \underline{\textit{minCorr}}: By increasing the value to 0.35, the arrow between \texttt{Review Score\_shift-1} is not kept anymore because the correlation between \texttt{Review Score\_shift-1} and \texttt{Publication\_Status\_shift-1} is below that value. Above 0.35, no DRD is discovered. When decreasing the value to 0.1, Figure \ref{PMIND Book DRD} is discovered as smaller correlation between variables are allowed. 
    \item \underline{\textit{minDev}}: determines how different models can be linked together by deciding how many object traces these should have in common. If the other hyperparameters are kept to 0.3, changing this hyperparameter does not change anything to the discovered DRD. This makes sense given that the hyperparameter values have a pretty high value. If the other hyperparameters such as \underline{\textit{minShift}} and \underline{\textit{minTraceprop}}  were to allow smaller clusters of object traces to be analyzed, than the \underline{\textit{minDev}} parameter will allow models to be build valid for a percentage of \% objects of that object cluster more than \underline{\textit{minDev}}.
    \item \underline{\textit{minSupport}}: By increasing \underline{\textit{minSupport}} to 0.9, no sufficiently strong predictive model is found. Reducing the value will not change the discovered models given that the other hyperparameters already remove smaller object clusters. Given that we evaluate on an artificial dataset, the rules apply to all object traces in any case therefore if one wants to find rules valid for smaller object clusters one would have to lower the values of \underline{\textit{minShift}} and \underline{\textit{minTraceprop}}.
    As a general rule given that \underline{\textit{minCorr}} selects individual variables based on their correlation and given that \underline{\textit{minSupport}} looks at the possible combinations of all the selected variables, setting \underline{\textit{minSupport}} higher than \underline{\textit{minCorr}} will select models with stronger predictive power and probably also fewer variables than the ones selected with \underline{\textit{minCorr}}. Setting \underline{\textit{minSupport}} lower than \underline{\textit{minCorr}} will generate other models with  smaller predictive power but might provide also with interesting insights for the process owners.
\end{itemize}

One of the key advantages of this algorithm is its procedural nature making it more interpretable than a pure black-box model.
Practically, IODDA is valuable in uncovering the timing and nature of decisions within various object-centric processes.
It effectively constructs a timeline, revealing when decisions were made.
In case an object-centric process log is generated with a consistent object-event mapping hence potentially generating less variation, the IODDA algorithm draws benefit from this by analyzing by object type making inference between object types more manageable and scalable. 
IODDA can be considered a process discovery algorithm, specifically focusing on activities that introduce or alter variables.
Whenever the ground-truth of the discovered decisions are available, it is possible to perform conformance analysis to see whether the prescribed decision logic is in line with what is discovered in the object-centric process log.
Even when the ground truth is not available, it can be valuable to discover which decisions are being taken within a process and what the logic of these decisions are.

In order to disentangle the discovered decision logic, the process owner or domain expert can further validate the results. 
As such, IODDA can be used by knowledge experts to confirm what should be happening in a process or can be used to discover decision logic that should maybe not be allowed within the process.

Finally, in the related work section an algorithm to discover DMN models from traditional XES event logs is mentioned \cite{de2019holistic}. Even though both algorithms approach the DMN model discovery problem in a similar vein, IODDA approaches it from a multi-dimensional perspective by incorporating the existence of multiple object types within the core of its algorithm which is not the case for the algorithm proposed in \cite{de2019holistic}. Therefore, analyzing a flattened DOCEL log with the algorithm in \cite{de2019holistic} will provide less insights than if the DOCEL log were to be directly analyzed with IODDA. This can mainly be attributed to the dimension reduction taking place during the flattening process where crucial information and relations might get lost which are simply not (re)-discoverable in a flattened event log. For example, in Figure \ref{PMIND Ship DRD}, IODDA is able to discover that the final decision depends on attributes from two different object types namely \texttt{Customers} and \texttt{Orders}. However if the object-centric event log were to be flattened to a traditional event log one would only be able to analyze the process through the lens of the \texttt{Customers} or the \texttt{Orders}, but because of the dimension reduction not all attributes might be correctly attributed or even kept which makes that the algorithm described in \cite{de2019holistic} simply does not have access to all the information.

\section{Limitations and Future Work} \label{Limitations and Future Work}

The algorithm currently supports DOCEL logs, with plans for future support for OCEL logs. The main challenge regarding OCEL logs, as explained previously, is to know which attributes belong to which object  \cite{goossens2023enhancing}. Because this object-attribute relation is a crucial foundation of the algorithm it is not trivial to adapt it to OCEL logs. 
Next, the algorithm was developed to discover activities that execute decisions as well as the decision logic, but currently does not discover decision points. This is because control flow information is not explicitly taken into account, but IODDA rather implicitly has control flow information such as XOR-splits through the analysis of decisions.
Supporting this in the future would provide an even more holistic overview of the executed decisions and how their outcome impact the control flow of the process. 
Also IODDA does not directly link the discovered DMN models to the process models but since IODDA identifies the activities responsible for executing decisions, this is certainly also possible.
Next, despite the algorithm being tested on two different realistic artificial logs, it would be beneficial to test the algorithm on a real-life object-centric event log. Unfortunately, due to a lack of publicly available object-centric logs, both in a DOCEL and OCEL format, in which decisions take place, this was not possible within this study. But given that OCEL logs can be converted into DOCEL log formats, we hope that public KiP OCEL logs will be made available to further test the algorithm on real-life object-centric logs.

Complex values such as lists or dictionaries currently form a challenge for the algorithm.
For example, imagine we have three different product types \texttt{\{P1,P2,P3\}} with \texttt{Value}: \{100,200,150\} and a customer enters an order with the following quantity for \texttt{Quantity:\{1,0,3\}}.
The first question is then, which \texttt{Quantity} and \texttt{Value} value refers to which product type as this is not explicitly provided together with the log but is rather implicit information.
Secondly, even if this information is known, one would still need to decide on a fitting correlation metric as it might be that there is no one size fits all metric as one could give equal or different weights related to the number of elements of the complex values however this only works if these are of equal lengths. In short, the support of complex values requires more research. In essence this issue can be solved by further normalizing the log but this has not been proposed in the object centric process literature yet and would require more investigation of its impact. 

Because event logs are an enactment of processes, not all decision paths are necessarily traversed in a sufficient amount to have sufficient (if any) correlation. As such, the complete ground truth DMN model can not always be discovered in full but only the paths which are executed within the log. Following up on the previous observation, because the initial variable selection procedure is based on the mutual information between two variables, complex relationships involving multiple variables might not always be discovered completely because of the fact that these two variables do not have enough correlation between them despite them being part of a complex relationship. Further investigation of a variable selection procedure addressing this aspect is also planned in the future.

\section{Conclusion} \label{Conclusion}

Object-centric event logs have been introduced to store business processes with multiple object types.
This development holds particular significance for decision mining, as organizations frequently make decisions by integrating information from various process-involved sources. Consequently, decisions can be inherently regarded as object-centric.
This paper proposes the first Integrated Object-centric  and Decision Discovery Algorithm (IODDA). IODDA is able to discover which activities introduce relevant decision variables and which activities execute decisions. Next to that, it is also able to relate the variables to the correct object types. Lastly, the algorithm is also able to discover the decision logic behind each decision. All together, a holistic overview of the process as well as a complete DMN model, containing both the discovered decision structure and decision logic, are discovered by the algorithm. The algorithm is implemented on a set of data-aware object-centric event logs which are made available to the public together with the first DOCEL log generators. Based on the results, it can be concluded that object-centric event logs can be used to automatically discover DMN models.

\section*{Acknowledgements}
This study was supported by the Research Foundation Flanders under grant numbers G079519N and G039923N and Internal Funds KU Leuven under grant number C14/23/031.

\bibliographystyle{elsarticle-num-names-alpha}
\bibliography{mybibliography}

\end{document}